%% file: dfu2022_galdran.tex
\documentclass[runningheads]{llncs}
\usepackage{graphicx}
\usepackage[english]{babel}
\usepackage[utf8]{inputenc}
\usepackage[caption=false,font=footnotesize]{subfig}
\usepackage[export]{adjustbox}
\usepackage{caption}
\usepackage{booktabs}
\usepackage{amsmath,amssymb}
\usepackage[pagebackref=true]{hyperref} 
\renewcommand*\backref[1]{\ifx#1\relax \else (Cited on #1) \fi}
\usepackage[table]{xcolor}
\usepackage{collcell}
\usepackage{hhline}
\usepackage{pgf}
\usepackage{multirow}
\usepackage{hyperref}
\usepackage{blindtext}
\usepackage{enumitem}
\usepackage{bbm}
\usepackage{siunitx}
\usepackage{arydshln}

\usepackage{accents}

\usepackage{wrapfig}
\usepackage{soul}
\usepackage[normalem]{ulem}
\usepackage{makecell}
\def\colorModel{hsb} %
\newcommand\ColCell[1]{
  \pgfmathparse{#1<50?1:0}  %
    \ifnum\pgfmathresult=0\relax\color{white}\fi
  \pgfmathsetmacro\compA{0}      %
  \pgfmathsetmacro\compB{#1/100} %
  \pgfmathsetmacro\compC{1}      %
  \edef\x{\noexpand\centering\noexpand\cellcolor[\colorModel]{\compA,\compB,\compC}}\x #1
  } 
\newcolumntype{E}{>{\collectcell\ColCell}m{0.45cm}<{\endcollectcell}}  %

\usepackage[misc,geometry]{ifsym}
\newcommand{\bceL}{$\mathcal{L}_{\mathrm{BCE}}$}
\newcommand{\bceplusdiceL}{$\mathcal{L}_{\mathrm{BCE}+\mathrm{Dice}}$}
\newcommand{\bcelindiceL}{$\mathcal{L}_{\mathrm{BCE} \rightsquigarrow \mathrm{Dice}}$}
\newcommand{\bceftdiceL}{$\mathcal{L}_{\mathrm{BCE} \rightarrow \mathrm{Dice}}$}
\newcommand{\diceL}{$\mathcal{L}_{\mathrm{Dice}}$}
\newcommand{\unl}[1]{\underline{#1}}

\begin{document}

\title{On the Optimal Combination of Cross-Entropy and Soft Dice Losses for Lesion Segmentation with Out-of-Distribution Robustness}

\titlerunning{Combination of CE \& Dice for Lesion Segmentation with OoD Robustness}
\author{Adrian Galdran\inst{1,2}$^\textrm{,\Letter}$ \and
Gustavo Carneiro\inst{2} \and
Miguel A. González Ballester\inst{1,3}}
\authorrunning{A. Galdran et al.}
\institute{BCN Medtech, Dept. of Information and Communication Technologies, Universitat Pompeu Fabra, Barcelona, Spain, \email{\{adrian.galdran,ma.gonzalez\}@upf.edu}
\and
University of Adelaide, Adelaide, Australia, \email{gustavo.carneiro@adelaide.edu}
\and
Catalan Institution for Research and Advanced Studies (ICREA), Barcelona, Spain }

\maketitle              %
\begin{abstract}
We study the impact of different loss functions on lesion segmentation from medical images. 
Although the Cross-Entropy (CE) loss is the most popular option when dealing with natural images, for biomedical image segmentation the soft Dice loss is often preferred due to its ability to handle imbalanced scenarios. 
On the other hand, the combination of both functions has also been successfully applied in this kind of tasks. 
A much less studied problem is the generalization ability of all these losses in the presence of Out-of-Distribution (OoD) data. 
This refers to samples appearing in test time that are drawn from a different distribution than training images.
In our case, we train our models on images that always contain lesions, but in test time we also have lesion-free samples. 
We analyze the impact of the minimization of different loss functions on in-distribution performance, but also its ability to generalize to OoD data, 
via comprehensive experiments on polyp segmentation from endoscopic images and ulcer segmentation from diabetic feet images.
Our findings are surprising: CE-Dice loss combinations that excel in segmenting in-distribution images have a poor performance when dealing with OoD data, which leads us to recommend the adoption of the CE loss for this kind of problems, due to its robustness and ability to generalize to OoD samples.
Code associated to our experiments can be found at \url{https://github.com/agaldran/lesion_losses_ood} .
\end{abstract}
\setcounter{footnote}{0} 
\section{Introduction}
In the field of medical image segmentation, lesion/background separation is a central image analysis task that shows up in many different applications, ranging from polyp detection from endoscopic frames \cite{ali_assessing_2022,hicks_endotect_2021} to wound segmentation \cite{kendrick_translating_2022,wang_fuseg_2022} or volumetric brain tumor segmentation \cite{islam_spatially_2021}. 
In all cases, the common setup is that we aim at partitioning the image domain into a foreground and a background region, with the foreground being typically smaller in size than the background.
This results in an class-imbalanced segmentation problem for which model optimization becomes harder than in a well-balanced scenario.

A key aspect of optimization in imbalanced problems is the choice of the loss function \cite{ma_loss_2021}. 
In natural image segmentation tasks the standard Cross-Entropy (CE) loss is the most commonly adopted cost function \cite{liu_hidden_2022}, whereas in medical image segmentation the soft Dice loss enjoys widespread popularity \cite{bertels_optimizing_2019}. 
The most likely reason for this is the typical imbalanced class distribution present in medical image segmentation: it has been shown that, for these problems, minimization of the soft Dice loss results in more accurate models, although it can induce potentially worse calibration \cite{mehrtash_confidence_2020}. 
Another widely adopted strategy is the combination of both CE and soft dice into a single loss function \cite{ma_loss_2021}. 
However, how to best combine each of them to achieve the reliability of CE \textit{plus} the ability of soft Dice to handle class-imbalanced problems remains unclear.

Calibration is the property by which model predictions reflect true probabilities, and thereby meaningful uncertainties, as opposed to over or underconfident outputs. 
Hence, model calibration is a critical aspect of performance for sensitive use-cases like fine localization of tumor borders \cite{mehrtash_confidence_2020,rousseau_post_2021}, but it is also important when dealing in test time with data that does not resemble the images used for model training. 
This sort of data is often referred to as Out-of-Distribution (OoD) \cite{hendrycks_many_2021,galdran_test_2022}, and it poses the challenge of how to learn models that can properly generalize to it. 
In this paper, we consider a specific but very relevant instance of an OoD generalization problem, namely models trained on images that always contain a lesion and need to handle in test time samples that may not contain a lesion but only background pixels. 
Our main contributions follow next.

\subsection*{Contributions and Paper Organization}
Our main technical contributions are:
\begin{itemize}[leftmargin=*]
    \item We empirically study which combinations of CE and Dice losses work best in two lesion segmentation problems, for two different scenarios: a) when we assume that a lesion is always present on the image both in the training and in the test set, and b) when this assumption does not hold for the test set.
    \item We clarify ambiguous definitions in the literature about what is OoD data in binary segmentation and frame lesion segmentation on images that can potentially not contain foreground as an OoD generalization task.
    \item This paper also serves as a detailed description of our submission to the Diabetic Foot Ulcer Challenge 2022, held in conjunction with MICCAI 2022.
\end{itemize}
The rest of the paper is organized as follows: we first introduce notation to describe the CE and the soft dice losses in the context of binary segmentation, with different strategies to combine them into a single loss function. 
We then reflect on the definition of Out-of-Distribution data in binary image segmentation, and argue about the formulation the no-foreground case as an OoD generalization task. 
This section is closed with the details of how we train segmentation models in this paper.
We then move to the next section, where we provide a description of the data that we employ for our experiments. 
Following, we report comprehensive experimental results and discuss the performance of each loss function the two considered segmentation problems.
We finish the paper with some conclusions and general recommendations that derive from our analysis. 

\section{Methodology}
In this section we first recall the formulation of the two loss functions used in this paper, and then introduce several simple loss combination mechanisms. 
Next, we formulate a definition of Out-of-Distribution data in the binary image segmentation context, and briefly describe other model training details.

\subsection{The Cross-Entropy loss, the Dice loss, and their Combinations}
It is worth to introduce first some notation. In the general case, let us consider:
\begin{itemize}[leftmargin=*]
\item A neural network $\mathcal{U}_\theta$ taking image $\mathbf{x}=\{x_1,\ldots,x_{|\Omega|}\}$ defined on a spatial domain $\Omega$, and generating a candidate segmentation $\mathcal{U}_\theta(\mathbf{x})=\hat{\mathbf{y}}=\{\hat{y}_1, \ldots, \hat{y}_{|\Omega|}\}$. 
\item A pixel $x_i$ belongs to one of $C$ possible categories, $\hat{y}_i=(\hat{y}_i^1,\ldots,\hat{y}_i^C) \in [0,1]^C$, and we typically apply a softmax layer to the output of $\mathcal{U}_\theta$ so that $\sum_{c=1}^C \hat{y}_i^c =1$, describing the probability of $x_i$ being from of each class. 
\item In addition, we have a ground-truth image $\mathbf{y}=\{y_1,\ldots, y_{|\Omega|}\}$, where each pixel has been annotated with a single integer value, the label of the correct category: $y_i\in\{1,\ldots, C\}$. 
\item Finally, from the prediction $\hat{\mathbf{y}}$ we are eventually interested in generating a segmentation that contains also integer values, which we will denote by $\bar{\mathbf{y}}$. This is often achieved by applying the arg-max operation on each vector $\hat{y}_i$.
\end{itemize}

We can therefore perform a pixel-wise comparison between $\mathbf{y}$ and $\mathbf{\hat{y}}$ by means of a loss function, and iteratively update the parameters $\theta$ of our model so that the loss is minimized.

\subsubsection{The Cross-Entropy Loss}
A popular loss function for (pixel-wise) classification problems is the Log-loss, also known as \textit{Cross-Entropy loss}, defined as:
\begin{equation}\label{ce}
\mathcal{L}_{CE} (\mathbf{y}, \mathbf{\hat{y}}) = -\frac{1}{|\Omega|} \sum_{i=1}^{|\Omega|} \sum_{c=1}^C  \mathbbm{1}^c(y_i) \cdot \log{\hat{y}}_i^c
\end{equation}
where $\mathbbm{1}^c(y_i)$ is zero unless pixel $x_i$ belongs to class $c$, \textit{i.e.} $y_i=c$, in which case $\mathbbm{1}^c(y_i)=1$. 
In other words, at a given pixel $x_i$ we look at its log-probability of belonging to the correct class, and average over all pixels on the image.

For the particular case of binary segmentation problems - \textit{foreground} versus \textit{background}, which is the focus of this paper, there are two alternative approaches. 
We can just set $C=2$ in the above description, so that for each pixel we have $\hat{y}_i=(\hat{y}_i^1,\hat{y}_i^2)$, but we can also realize that due to the constraint of probabilities summing up to one, $\hat{y}_i^2 = 1-\hat{y}_i^1$. 
Therefore we could also keep only $\hat{y}_i^2$, or replace the softmax layer by a sigmoid operation that produces a single number in the $[0,1]$ 
interval\footnote{In this case, the mechanism to build a segmentation $\bar{\mathbf{y}}$ by taking the argmax over $\hat{\mathbf{y}}$ would be equivalent to thresholding the sigmoid output with a value of $t=0.5$.}, 
and have the annotations be either $0$ or $1$, $y_i\in\{0,1\}$. 
Simply speaking, the model now provides a single output $\hat{y}_i$ per pixel representing the probability of belonging to the foreground class. 
In this case, the \textit{Binary Cross-Entropy} loss can be written as:
\begin{equation}\label{bce}
\mathcal{L}_{BCE} (\mathbf{y}, \mathbf{\hat{y}}) = -\frac{1}{|\Omega|} \sum_{i=1}^{|\Omega|} y_i \cdot \log(\hat{y}_i) + (1-y_i) \cdot \log(1-\hat{y}_i).
\end{equation}
For foreground pixels ($y_i=1$) only the predicted log-probability $\log(\hat{y}_i)$ of belonging to the foreground contributes to the loss, while for background pixels ($y_i=0$) the contribution is the log-probability of belonging to the background, $\log(1-\hat{y}_i)$. 
Therefore, equations (\ref{ce}) and (\ref{bce}) are fully equivalent.

\subsubsection{The Soft-Dice Loss}
Another popular segmentation loss is the soft Dice, introduced for the binary case in \cite{milletari_v-net_2016} and later generalized to a (frequency-weighted) multi-class setting in \cite{sudre_generalised_2017}. 
It is a differentiable reformulation of the well-known Dice Similarity Score (DSC), a measure of set similarity given by:
\begin{equation}
\textrm{DSC}(X,Y) = \frac{2\,|X \cap Y|}{|X| + |Y|}
\end{equation}
where the operator $|\cdot|$ represents a count of the number of elements in the set. 
If $X$ and $Y$ overlap perfectly, $\textrm{DSC}(X,Y)=1$, and when the overlap decreases, then $|X \cap Y|$ diminishes while $|X| + |Y|$ is preserved, reducing the score.

In the binary segmentation case, we can use DSC to assess the performance of our model by measuring the similarity of the manual ground-truth $\mathbf{y}$ and a segmentation $\bar{\mathbf{y}}$. 
This can be done by defining $\mathbf{y} \cap \mathbf{\bar{y}} = \{i \in \Omega \ | \ y_i = \bar{y}_i = 1\}$ and $\mathbf{y} \cup \mathbf{\bar{y}} = \{i \in \Omega \ | \ y_i=1 \textrm{ or } \bar{y}_i=1\}$. 
In addition, we can write these quantities in terms of True/False Positives (TP, FP) and True/False Negatives (TN/FN) as $\mathbf{\hat{y}} \cap \mathbf{y} = \mathrm{TP}$ and $\mathbf{\hat{y}} \cup \mathbf{y} = 2\mathrm{TP} + \mathrm{FP} + \mathrm{FN}$, \textit{i.e.}:
\begin{equation}\label{dsc}
\textrm{DSC}(\mathbf{y},\mathbf{\bar{y}}) = \frac{2|\mathbf{y} \cap \mathbf{\bar{y}}|}{|\mathbf{y}| + |\mathbf{\bar{y}}|} = \frac{2 \mathrm{TP}}{2\mathrm{TP} + \mathrm{FP} + \mathrm{FN} },
\end{equation}
which shows that the Dice Similarity Score disregards True Negatives. 
This can be an advantage whenever TNs compose the majority of pixels on a segmentation, which is often the case in medical applications where large background regions are easily predicted by a model.

In order to turn eq. (\ref{dsc}) into a loss function, we need it to take a binary ground-truth $\mathbf{y}$ and a continuous prediction $\mathbf{\hat{y}}$, and return a quantity that is differentiable, and decreases as the prediction improves \cite{eelbode_optimization_2020}. 
This can be achieved by realizing that $|\mathbf{y} \cap \mathbf{\hat{y}}| = \sum_i y_i \cdot \hat{y}_i = \langle \mathbf{y}, \mathbf{\hat{y}}\rangle$, and $|\mathbf{y}| = \sum_i y_i \cdot y_i = \langle \mathbf{y}, \mathbf{y}\rangle$, so that we can write:
\begin{equation}\label{dice}
\mathcal{L}_{Dice}(\mathbf{y}, \mathbf{\hat{y}}) = 1 - \frac{2|\mathbf{y} \cap \mathbf{\hat{y}}|}{|\mathbf{y}| + |\mathbf{\hat{y}}|} = 1-\frac{2\langle \mathbf{y}, \mathbf{\hat{y}}\rangle}{\langle \mathbf{y}, \mathbf{y}\rangle + \langle \mathbf{\hat{y}}, \mathbf{\hat{y}}\rangle}.
\end{equation}

Finally, let us remark that, as opposed to the Cross-Entropy loss, the Dice loss is not defined at the pixel level but at the image level, and it degenerates if the ground-truth does not contain any foreground pixel, i.e.  $\mathbf{y}=\mathbf{0}$. 
In this case, it will always be maximal regardless of the prediction $\mathbf{\hat{y}}$, but as the prediction improves ($\mathbf{\hat{y}} \rightarrow\mathbf{0}$), the denominator approaches zero, which can result in numerical instabilities during training.

\subsubsection{Loss Combinations}

The relationship between the CE and the soft Dice losses, and how to optimally combine them, has recently drawn attention in the research community both from a theoretical \cite{liu_hidden_2022,yeung_unified_2022} and from a practical perspective \cite{ma_loss_2021}. 
It is common belief that the soft Dice loss can perform better in highly imbalanced scenarios and can result in DSC improvements \cite{bertels_optimizing_2019}, although it is known that this may come at the cost of poor calibration \cite{mehrtash_confidence_2020}, which may require specific training techniques \cite{islam_spatially_2021,gros_softseg_2021,liu_devil_2022} or post-processing methods or fine-tuning \cite{rousseau_post_2021}. 

In order to leverage the best of both worlds, one can also consider combining the two losses during training \cite{ma_loss_2021}. 
Indeed, even works that advocate for the use of the soft Dice loss over BCE, like \cite{eelbode_optimization_2020} start the training with a preliminary stage in which models are trained until convergence with CE, and only then they are fine-tuned using soft Dice. 
In this work, we adopt five different loss combination strategies, namely:
\begin{enumerate}%
    \item \textbf{Only Binary Cross-Entropy.} We minimize $ \mathcal{L}_{\mathrm{BCE}}(\mathbf{y}, \mathbf{\hat{y}})$ alone.
    \item \textbf{Loss addition}. A simple combination with equal weights on both losses: 
    \begin{equation}\label{loss_add}
        \mathcal{L}_{\mathrm{BCE}+\mathrm{Dice}}(\mathbf{y}, \mathbf{\hat{y}}) = \mathcal{L}_{\mathrm{BCE}}(\mathbf{y}, \mathbf{\hat{y}}) + \mathcal{L}_{\mathrm{Dice}}(\mathbf{y}, \mathbf{\hat{y}}).
    \end{equation}
    \item \textbf{Soft Fine-tuning}. We minimize a linear combination that starts giving full weight to BCE and ends up giving only weight to Dice, with intermediate weights linearly interpolated. At epoch $n=0, 1, \ldots N$ this loss is given by:
    \begin{equation}\label{soft_ft}
    \mathcal{L}_{\mathrm{BCE} \rightsquigarrow \mathrm{Dice}}(\mathbf{y}, \mathbf{\hat{y}}) = \frac{(N-n)}{N} \cdot \mathcal{L}_{\mathrm{BCE}}(\mathbf{y}, \mathbf{\hat{y}}) + \frac{n}{N} \cdot \mathcal{L}_{\mathrm{Dice}}(\mathbf{y}, \mathbf{\hat{y}})
    \end{equation}
    \item \textbf{Hard Fine-tuning}. We minimize the BCE loss in the first part of the training and switch to the Dice loss only in the last 10\% of the training. 
    For a model that trains over $N$ epochs, at epoch $n$ this is:
    \begin{equation}\label{hard_ft}
     \mathcal{L}_{\mathrm{BCE} \rightarrow \mathrm{Dice}}(\mathbf{y}, \mathbf{\hat{y}})= 
     \begin{cases}
			\mathcal{L}_{BCE}(\mathbf{y}, \mathbf{\hat{y}}), & \text{if $n<0.9\cdot N$}\\
            \mathcal{L}_{\mathrm{Dice}}(\mathbf{y}, \mathbf{\hat{y}}), & \text{otherwise}
		 \end{cases}   
    \end{equation}
    \item \textbf{Only soft Dice}. We minimize $ \mathcal{L}_{\mathrm{Dice}}(\mathbf{y}, \mathbf{\hat{y}})$ alone.
\end{enumerate}

\subsection{On the definition of In-Distribution and Out-of-Distribution Data for Binary Segmentation Problems}
The definition of what constitutes Out-of-Distribution (OoD) data is not clearly agreed in the machine learning community, let alone for binary segmentation tasks, so it is worth briefly clarifying the perspective we adopt here. 

For classification problems, sometimes OoD data is defined as data that does not belong to any of the categories on which the model was trained \cite{hendrycks_baseline_2017}, and the goal is to detect it and abstain from classifying it, a task known as \textbf{OoD Detection}. 
Other times, OoD data is considered as arising from some sort domain shift, \textit{e.g.} different acquisition conditions, but not from a semantic shift - this is, OoD data belongs to the same set of categories as the in-distribution training data \cite{tran_plex_2022}. 
From this point of view, our model should be able to generalize to out-of-distribution data and classify correctly, a property known as \textbf{Robustness to OoD} in this context \cite{hendrycks_many_2021}.

Image segmentation can be considered as a per-pixel classification problem, and in this case one can directly apply the above definition of OoD detection. 
The goal would be finding pixels on a test image that belong to anomalies not present in the training set \cite{popescu_distributional_2021}. 
A particularly popular approach to solve this problem is reconstruction-based models \cite{baur_autoencoders_2021}, in which an auto-encoder is tasked with rebuilding the input after reducing its dimensionality, and we expect it to fail in reconstructing OoD data. 
Since we can measure the reconstruction error pixel-wise, we are then able to localize OoD pixels individually. 
These could correspond to, for example, different pathologies like tumors or lesions, but also image quality degradations like localized visual artifacts \cite{zimmerer_mood_2022}.

On the other hand, in this paper we consider a more subtle OoD scenario that only exists for binary segmentation models, which are trained to receive an image $\mathbf{x}$ and produce a partition of its support $\Omega$ into foreground and background regions $\Omega^f$ and $\Omega^b$.
Specifically, we are interested in analyzing the performance of one such model on an in-distribution test set that contains the same kind of foreground objects, but also the degradation of model performance when test images \textit{do not contain any foreground pixel}, this is, $\Omega=\Omega^b$. 

\subsection{Model Training Details}
All the models in this paper were trained following our previous work on lesion segmentation \cite{galdran_double_2021}, which we have applied successfully in several other biomedical segmentation challenges \cite{hicks_endotect_2021,ali_assessing_2022,wang_fuseg_2022}. 
Specifically, our architectures are always a cascade of two encoder-decoder networks, the encoder being pretrained on ImageNet and the decoder being a Feature-Pyramid Network \cite{lin_feature_2017}, and we consider several popular encoder architectures with increasing sizes. 
We optimize network weights by minimizing different loss combinations as described above, using an Adam optimizer with a learning rate of $l=3e{\text -}4$ and a batch-size of $4$. 
The learning rate is decayed to zero following a cosine law during training, which lasts for 40,000 optimization steps, which we empirically found enough for model convergence. 
Images are resized to $640\times512$, which was the most common resolution in both of the considered datasets, and augmented during training with conventional image processing operations (random rotations, translations, scalings, vertical/horizontal flipping, contrast/saturation/brightness changes, etc.). 
We do not apply early stopping and simply keep the final model weights for testing purposes.

\section{Experimental Results}
We describe in this section our experimental validation on lesion segmentation in presence of no-foreground OoD data.

\subsection{Data and Performance Measures}
Data in this paper comes from two different sources\footnote{DFUC 2022 Challenge: \url{https://dfuc2022.grand-challenge.org/} \\Endotect 2020 Challenge: \url{https://endotect.com/}}: the Diabetic Foot Ulcer Segmentation challenge DFUC 2022 \cite{kendrick_translating_2022} and the Endotect 2021 challenge \cite{hicks_endotect_2021}. 

\begin{figure}[!t]
    \centering
    \subfloat[]{\includegraphics[width = 0.24\textwidth,valign=c]{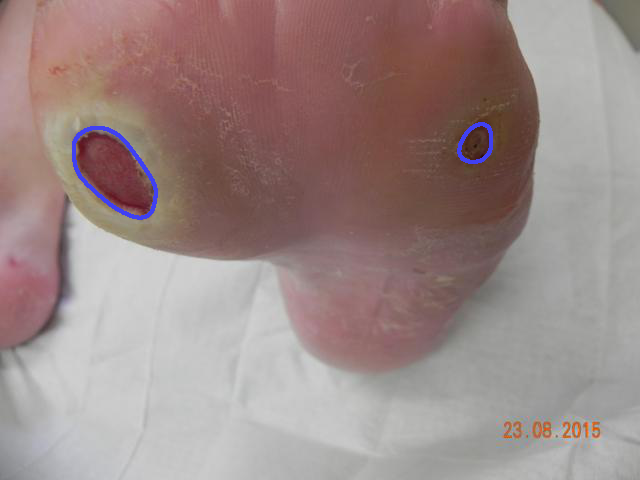}
    \label{fig_dfu1}}
    \hfil
    \subfloat[]{\includegraphics[width = 0.24\textwidth,valign=c]{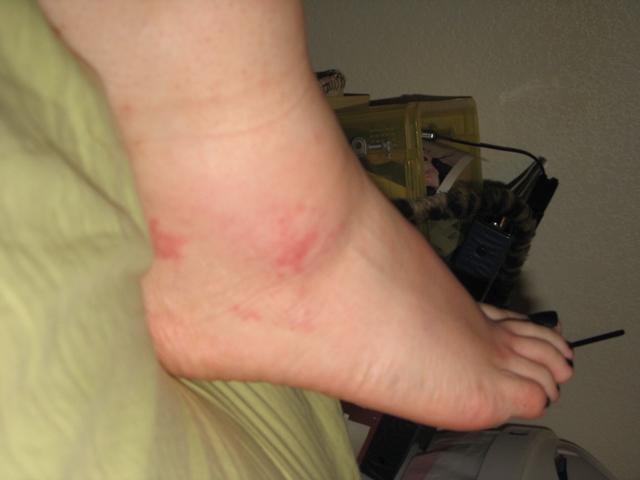}
    \label{fig_polyp1}}
    \hfil
    \subfloat[]{\includegraphics[width = 0.24\textwidth,valign=c]{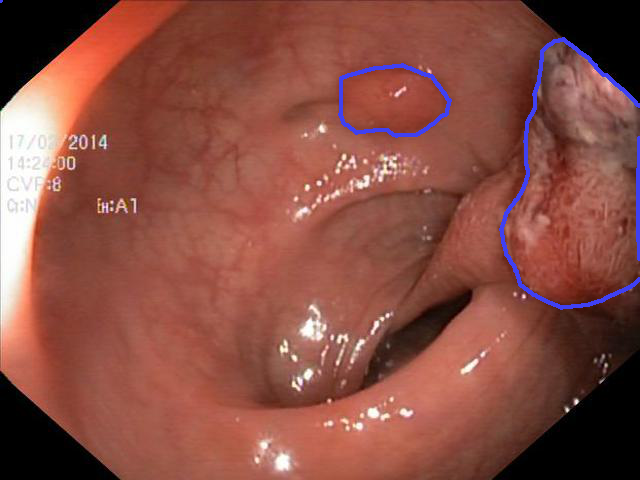}
    \label{fig_dfu2}}
    \hfil
    \subfloat[]{\includegraphics[width = 0.24\textwidth,valign=c]{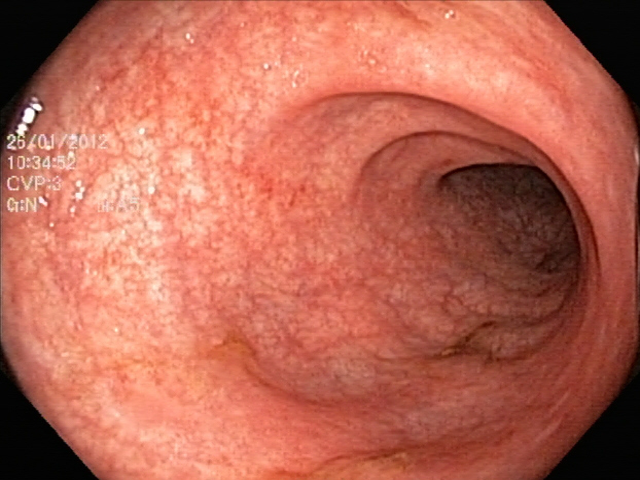}
    \label{fig_polyp2}}
    \caption{The images in (a) and (c) contain a lesion marked in blue; (a) foot ulcer, (c) polyp. 
    The other two images contain no foreground: (b) DFU remission case, (d) ulcerative colitis. 
    We treat the latter as OoD data in this work.}
    \label{examples}
\end{figure}

\begin{itemize}[leftmargin=*]
\item The DFUC 2022 dataset contains 4,000 $640\times480$ images divided equally into a training and a test set\cite{kendrick_translating_2022}. 
The test set contains some cases of healthy individuals which are useful for testing our models in the absence of lesions. 
At the time of writing we do not have access to the test set, but the organizers have made available a small subset with 200 images, which also contains no-foreground samples.
Submission to the online grand-challenge website returns average performances on this test set, which we report below.
\item The Endotect 2021 dataset contained has 10,662 labeled endoscopic images capturing 23 different classes of findings, extracted from the Hyper-Kvasir dataset \cite{borgli_hyperkvasir_2020}. 
There are also segmentation masks for 1,000 images from the polyp class, which we use to learn binary segmentation models. 
In test time, we use the polyp segmentation test set from the Endotect challenge, which contains 200 images containing polyps and corresponding binary masks, and we expand this set with another 20 images randomly sampled from the ulcerative collitis category in the Endotect dataset, \textit{i.e.} images that do not contain a polyp but other kind of pathology. 
Fig. \ref{examples} shows some examples of polyp and no-lesion cases, as well as examples from the DFUC 2022 dataset.\end{itemize}

In this paper we use a five-fold cross-validation approach: the training set is split in five subsets of equal size, one of them serving as a validation set. 
The validation set is rotated and training takes places five times, resulting in five sets of trained weights that we use for average performance estimation, and also for ensembling when generating predictions on the test set.

Performance is measured in terms of Dice Score (DSC) as in eq. (\ref{dsc}). 
Note that, following the DFUC 2022 guidelines, we assign a score of 0 to predictions that contain a lesion when the ground-truth does not, and to predictions that contain no lesion when the ground-truth shows a lesion. 
A score of 1 is assigned to the case in which neither the ground-truth nor the prediction contain a lesion.

\subsection{Performance Evaluation and Discussion}
\subsubsection{Performance on DFU Segmentation}
\input{table_dfu_ID}
We show in Table \ref{dfu_id} results of 5-fold cross-validation experiments for DFU segmentation on \textbf{in-distribution data}, \textit{i.e.} images that always contain a lesion. 
In this and the remaining cases of this section, we report average DSC for a range of CNN architectures, sorted top-to-bottom from smaller to larger in terms of learnable weights. 

By looking at Table \ref{dfu_id} we quickly realize that the training with \bceplusdiceL \ and \ \bcelindiceL \ invariably leads to models achieving a stronger performance in terms of average DSC, regardless of the architecture. 
Indeed, any combination of the two base loss functions results on an improved performance when compared with using either \bceL \ or  \ \diceL \ alone for driving the optimization.
It is also worth noting that training with \bceL \ leads to the worst results in almost all cases, unless for the largest encoders ResNet152 and ResNext101. 

Results are strikingly different when it comes to testing on a mixture of in-distribution and \textbf{OoD data}, \textit{i.e.} images that may not contain a lesion. 
Table \ref{dfu_ood} shows the average DSCs for the same set of models after 5-fold ensembling and submission to the DFU Challenge 2022 website, and we can see that the above trend is fully reverted. 
The weakest loss function, \bceL, becomes the strongest one in the presence of OoD images, largely outperforming the previously two best approaches, unless for the two smallest models. 
In addition, the Hard Fine-tuning approach using \bceftdiceL , which was the loss combination with the lowest performance before, is now one of the most reliable solutions, achieving the best performance in three out of the nine considered models, and reaching the second position in another four cases. 
The poor performance of the \diceL \ loss when optimized alone is also remarkable: irrespective of the size of the underlying architecture, the resulting DSC remains the worst of all considered approaches.

\input{table_dfu_OOD}

\subsubsection{Performance on Polyp Segmentation}
In the polyp segmentation problem we see a similar trend as for DFU segmentation. 
Namely, approaches that perform best for in-distribution data become again much worse when dealing with lesion-free images. 
As we can see in Table \ref{endo_id}, training with \bcelindiceL \ and \ \bceplusdiceL \ always results in the best performance, no matter the architecture. 
Perhaps the major difference with the previous case is the better average DSC achieved by \bceL, which now ranks as the third best option for most models. 
When we turn to testing on a combination of in-distribution and OoD data, we see again that both the Hard Fine-Tuning approach and the \bceL \ loss, together with the \bcelindiceL \ loss this time, achieve the highest average DSC, whereas the very popular \bceplusdiceL \ combination result in poor performance in most cases. 
Here again the optimization of \diceL \ leads to the worst results across the board.

\input{table_endo_ID}

\subsubsection{Submission to DFU Challenge}
In a last stage of the competition, the organizers requested submission on a large test set which included 2000 images, again with OoD data. 
We submitted the result of a five-fold ensembling of our best model (ResNeXt-101) with Test-Time Augmentation. 
We also used a pseudo-labeling technique like the one described in \cite{galdran_state---art_2022} to improve generalization ability. 
Our final model reached the third place with DSC$=72.63$, closely following the performance of the winner participant with a DSC$=72.87$.

\section{Conclusion}
While it is well-known that adding \bceL \ and \diceL \ results in stronger in-distribution segmentation performance, the behavior of \bceplusdiceL \ on OoD data is unexplored. 
Our findings show that in that scenario it is not be the best option due to poor robustness to the absence of lesions, and the Hard Fine-Tuning approach or even the standard \bceL \ loss may be more reliable. 
Note however that the use of \bceftdiceL \ requires setting a hyper-parameter: the number of epochs needed in order to reach convergence. 
Finally, although it is known that \diceL \ is poorly calibrated \cite{mehrtash_confidence_2020}, it is also common knowledge that ensembling models results in improved calibration. 
For that reason, the performance of \diceL \ in our OoD experiments is surprisingly disappointing, and we recommend avoiding the optimization of the soft Dice loss function alone in these cases.

\input{table_endo_OOD}

\section*{Acknowledgments}
This work was supported by a Marie Skłodowska-Curie Fellowship (No 892297) and by Australian Research Council grants (DP180103232 and FT190100525).

\bibliographystyle{splncs04}
\bibliography{seg_dfu.bib}

\end{document}

%% file: table_dfu_ID.tex
\begin{table}[!b]
\renewcommand{\arraystretch}{1.25}	
\setlength\tabcolsep{2.5pt}	
\begin{center}
\begin{tabular}{ccccccc}
\textbf{Enc./Loss}   &       \bceL        &    \diceL           &  |  &     \bceftdiceL     &  \bcelindiceL                    &   \bceplusdiceL               \\
\midrule
\textbf{MobileNet}    & 74.85$\,\pm\,$0.48 & 77.45$\,\pm\,$0.61  &  |  & 75.93$\,\pm\,$0.88  & \textbf{77.91}$\,\pm\,$0.61      & \unl{\textbf{77.94}}$\,\pm\,$0.45   \\
\midrule
\textbf{ResNet18}     & 74.73$\,\pm\,$0.99 & 76.90$\,\pm\,$0.67  &  |  & 76.20$\,\pm\,$0.71  & \textbf{77.57}$\,\pm\,$0.65      & \unl{\textbf{77.80}}$\,\pm\,$0.6    \\
\midrule
\textbf{ResNet34}     & 76.21$\,\pm\,$0.56 & 77.73$\,\pm\,$0.38  &  |  & 77.66$\,\pm\,$0.49  & \textbf{78.13}$\,\pm\,$0.39      & \unl{\textbf{78.69}}$\,\pm\,$0.42  \\
\midrule
\textbf{ResNet50}     & 76.37$\,\pm\,$0.45 & 76.94$\,\pm\,$0.49  &  |  & 77.04$\,\pm\,$0.32  & \textbf{78.02}$\,\pm\,$0.51      & \unl{\textbf{78.13}}$\,\pm\,$0.27 \\
\midrule
\textbf{ResNeXt50}    & 76.38$\,\pm\,$1.05 & 77.72$\,\pm\,$0.66  &  |  & 77.14$\,\pm\,$0.97  & \textbf{78.60}$\,\pm\,$0.67      & \unl{\textbf{78.82}}$\,\pm\,$0.46  \\
\midrule
\textbf{ResNet101}    & 76.55$\,\pm\,$1.06 & 76.89$\,\pm\,$0.70  &  |  & 77.36$\,\pm\,$0.36  & \textbf{78.04}$\,\pm\,$0.59      & \unl{\textbf{78.46}}$\,\pm\,$0.67  \\
\midrule
\textbf{ResNeXt101}   & 78.51$\,\pm\,$0.75 & 77.32$\,\pm\,$1.01  &  |  & 78.59$\,\pm\,$0.52  & \unl{\textbf{79.48}}$\,\pm\,$0.43 & \textbf{78.97}$\,\pm\,$0.40   \\
\midrule
\textbf{ResNet152}    & 77.07$\,\pm\,$0.36 & 76.73$\,\pm\,$0.52  &  |  & 77.95$\,\pm\,$0.36  & \unl{\textbf{78.52}}$\,\pm\,$0.67 & \textbf{78.30}$\,\pm\,$0.43  \\
\bottomrule
\\[-0.25cm]
\end{tabular}
\caption{Results with different architectures and loss functions for the task of \textbf{DFU segmentation}. 
For each model, \unl{\textbf{best}} and \textbf{second best} are marked. Performance is DSC averaged over 5-fold training, only \textbf{in-distribution} data.}\label{dfu_id}
\end{center}
\vspace{-1cm}
\end{table}

%% file: table_dfu_OOD.tex
\begin{table}[!t]
\renewcommand{\arraystretch}{1.25}	
\setlength\tabcolsep{1pt}	
\begin{center}
\begin{tabular}{ccccccc}
\textbf{Enc./Loss}   &       \bceL                 &    \diceL   &  |   &     \bceftdiceL             &  \bcelindiceL  &   \bceplusdiceL              \\
\midrule
\textbf{MobileNet}    & \textbf{65.73} (6.5\%)     & 65.24 (2\%)  &  |   & 65.35 (3.5\%)               & \ul{\textbf{66.25}} (1.5\%) & 65.56 (2\%)        \\
\midrule
\textbf{ResNet18}     & 66.05 (7\%)                & 64.47 (1\%)  &  |   & \textbf{66.21} (5\%)        & 66.10 (1\%)    & \ul{\textbf{66.57}} (1\%)               \\
\midrule
\textbf{ResNet34}     & \textbf{\ul{68.14}} (7\%)  & 64.84 (1\%)  &  |   & \textbf{67.60} (4.5\%)      & 66.53 (1.5\%) & 66.25 (1\%)                   \\
\midrule
\textbf{ResNet50}     & \textbf{68.03} (7\%)       & 65.21 (1\%)  &  |   & \unl{\textbf{68.26}} (4.5\%) & 65.87 (1.5\%) & 64.99 (0.5\%)                   \\
\midrule
\textbf{ResNeXt50}    & \textbf{67.65} (8\%)       & 64.85 (0\%)  &  |   & \unl{\textbf{67.99}} (4\%)   & 66.51 (1.5\%) & 66.87 (1.5\%)                  \\
\midrule
\textbf{ResNet101}    & \textbf{67.92} (8\%)       & 63.98 (1\%)  &  |   & \unl{\textbf{68.29}} (5\%)   & 65.09 (1\%) & 65.29 (2\%)                   \\
\midrule
\textbf{ResNeXt101}   & \unl{\textbf{68.70}} (5\%)  & 64.45 (0.5\%)  &  |  & \textbf{68.03} (4\%)       & 66.62 (1\%) & 65.49 (1\%)                   \\
\midrule
\textbf{ResNet152}    & \unl{\textbf{68.82}} (7\%)  & 65.09 (0.5\%)  &  |  & \textbf{67.63} (4.5\%)     & 66.21 (1\%) & 66.32 (1\%)                          \\
\bottomrule
\\[-0.25cm]
\end{tabular}
\caption{Results with different architectures and loss functions for the task of \textbf{DFU segmentation}. 
For each model, \unl{\textbf{best}} and \textbf{second best} are marked. Performance is DSC (\% rejected images) achieved by a 5-fold ensemble on the hidden validation set, which included \textbf{OoD data}.}\label{dfu_ood}
\end{center}
\vspace{-1cm}
\end{table}

%% file: table_endo_ID.tex
\begin{table}[!t]
\renewcommand{\arraystretch}{1.25}	
\setlength\tabcolsep{2.5pt}	
\begin{center}
\begin{tabular}{ccccccc}
\textbf{Enc./Loss}   &       \bceL        &    \diceL           &  |  &     \bceftdiceL     &  \bcelindiceL                    &   \bceplusdiceL               \\
\midrule
\textbf{MobileNet}    & 88.66$\,\pm\,$0.84 & 88.63$\,\pm\,$0.58  &  |  & 88.62$\,\pm\,$0.76  & \ul{\textbf{89.42}}$\,\pm\,$0.65 & \textbf{89.38}$\,\pm\,$0.50   \\
\midrule
\textbf{ResNet18}     & 89.25$\,\pm\,$1.49 & 88.99$\,\pm\,$1.07  &  |  & 89.28$\,\pm\,$1.50  & \ul{\textbf{89.96}}$\,\pm\,$1.00 & \textbf{89.55}$\,\pm\,$1.12    \\
\midrule
\textbf{ResNet34}     & 90.22$\,\pm\,$0.51 & 89.02$\,\pm\,$1.10  &  |  & 89.78$\,\pm\,$0.89  & \textbf{90.40}$\,\pm\,$0.98      & \ul{\textbf{90.59}}$\,\pm\,$0.57  \\
\midrule
\textbf{ResNet50}     & 90.36$\,\pm\,$1.25 & 88.82$\,\pm\,$0.70  &  |  & 90.13$\,\pm\,$0.91  & \unl{\textbf{90.71}}$\,\pm\,$0.90 & \textbf{90.67}$\,\pm\,$0.90  \\
\midrule
\textbf{ResNeXt50}    & 90.45$\,\pm\,$0.93 & 89.47$\,\pm\,$0.83  &  |  & 89.97$\,\pm\,$0.73  & \textbf{90.70}$\,\pm\,$0.83      & \unl{\textbf{90.84}}$\,\pm\,$0.74  \\
\midrule
\textbf{ResNet101}    & 90.49$\,\pm\,$0.90 & 88.36$\,\pm\,$0.62  &  |  & 90.30$\,\pm\,$0.79  & \textbf{90.88}$\,\pm\,$0.87      & \unl{\textbf{90.96}}$\,\pm\,$0.75  \\
\midrule
\textbf{ResNeXt101}   & 91.05$\,\pm\,$1.13 & 88.33$\,\pm\,$1.30  &  |  & 90.69$\,\pm\,$1.13  & \textbf{91.11}$\,\pm\,$0.79               & \unl{\textbf{91.18}}$\,\pm\,$0.73  \\
\midrule
\textbf{ResNet152}    & 90.47$\,\pm\,$0.69 & 88.83$\,\pm\,$1.35  &  |  & \textbf{90.78}$\,\pm\,$0.87  & \unl{\textbf{90.97}}$\,\pm\,$0.87               & 90.64$\,\pm\,$0.84  \\
\bottomrule
\\[-0.25cm]
\end{tabular}
\caption{Results with different architectures and loss functions for the task of \textbf{polyp segmentation}. 
For each model, \unl{\textbf{best}} and \textbf{second best} are marked. Performance is DSC averaged over 5-fold training, only \textbf{in-distribution} data.}\label{endo_id}
\end{center}
\vspace{-1cm}
\end{table}

%% file: table_endo_OOD.tex
\begin{table}[!t]
\renewcommand{\arraystretch}{1.25}	
\setlength\tabcolsep{2.5pt}	
\begin{center}
\begin{tabular}{ccccccc}
\textbf{Enc./Loss}    &       \bceL                   &    \diceL     &  |   & \bceftdiceL               &  \bcelindiceL             & \bceplusdiceL               \\
\midrule
\textbf{MobileNet}    & 86.64 (6\%)                   & 81.10 (1\%)   &  |   & \textbf{86.90} (6\%)      & \ul{\textbf{86.90}} (6\%) & 83.91 (3\%)                \\
\midrule
\textbf{ResNet18}     & \ul{\textbf{87.72}} (7\%)     & 80.66 (0\%)   &  |   & \textbf{87.19} (6\%)      & 86.65 (5\%)               & 85.17 (4\%)                \\
\midrule
\textbf{ResNet34}     & \textbf{88.40} (7\%)          & 82.89 (3\%)   &  |   & \ul{\textbf{89.78}} (8\%) & 87.35 (5\%)               & 85.78 (4\%)                \\
\midrule
\textbf{ResNet50}     & \textbf{88.40} (7\%)          & 81.20 (1\%)   &  |   & \unl{\textbf{89.67}} (8\%) & 88.30 (6\%)               & 85.22 (4\%)                \\
\midrule
\textbf{ResNeXt50}    & 88.85 (7\%)                   & 83.40 (2\%)   &  |   & \textbf{89.06} (7\%)      & \unl{\textbf{89.15}} (6\%) & 85.22 (3\%)                \\
\midrule
\textbf{ResNet101}    & \unl{\textbf{89.33}} (7\%)     & 81.36 (1\%)   &  |   & \textbf{89.12} (8\%)      & 88.15 (6\%)               & 87.92 (6\%)                \\
\midrule
\textbf{ResNeXt101}   & \textbf{90.41} (8\%)          & 81.54 (0\%)   &  |   & 90.26 (8\%)               & \unl{\textbf{90.48}} (8\%)      & 89.65 (7\%)                \\
\midrule
\textbf{ResNet152}    & \unl{\textbf{90.48}} (8\%)     & 81.66 (1\%)   &  |   & 89.77 (8\%)               & 88.66 (7\%)               & 88.67 (6\%)                \\
\bottomrule
\\[-0.25cm]
\end{tabular}
\caption{Results with different architectures and loss functions for the task of \textbf{polyp segmentation}. 
For each model, \unl{\textbf{best}} and \textbf{second best} are marked. Performance is DSC (\% rejected images, in this case there were 9\% OoD images) achieved by a 5-fold ensemble on the test set, which included \textbf{OoD data}.}\label{endo_ood}
\end{center}
\vspace{-1.01cm}
\end{table}